\documentclass[journal]{IEEEtran}
\usepackage{authblk}
\usepackage{graphicx}

\ifCLASSINFOpdf
\else
\fi
\begin{document}
%
\title{Black Box to White Box: Discover Model Characteristics Based on Strategic Probing}
%
%
%

\author[1, 2]{\small Josh Kalin}
\author[2]{\small Matthew Ciolino}
\author[2]{\small David Noever}
\author[1]{\small Gerry Dozier}
\affil[1]{\footnotesize Department of Computer Science and Software Engineering, Auburn University, Auburn, AL, USA}
\affil[2]{\footnotesize PeopleTec, Inc, 4901 Corporate Dr NW, Huntsville, AL 35805, USA}


%
%

\markboth{Ai4i 2020}%
{Shell \MakeLowercase{\textit{et al.}}: Bare Demo of IEEEtran.cls for IEEE Journals}
%



\maketitle

\begin{abstract}
In Machine Learning, White Box Adversarial Attacks rely on knowing underlying knowledge about the model attributes. This works focuses on discovering to distrinct pieces of model information: the underlying architecture and primary training dataset. With the process in this paper, a structured set of input probes and the output of the model become the training data for a deep classifier. Two subdomains in Machine Learning are explored - image based classifiers and text transformers with GPT-2. With image classification, the focus is on exploring commonly deployed architectures and datasets available in popular public libraries. Using a single transformer architecture with multiple levels of parameters, text generation is explored by fine tuning off different datasets. Each dataset explored in image and text are distinguishable from one another. Diversity in text transformer outputs implies further research is needed to successfully classify architecture attribution in text domain.
\end{abstract}


%
\IEEEpeerreviewmaketitle

\section{Introduction}
The goal is to classify the classifier. The approach presented in this work will show the ability of an attacker to discover model attributes about a model by simply understanding the distribution of the data and the outputs of common model architectures. This paper is presented in five sections: Introduction, Approach, Evaluation, Future Work, and Conclusions.

\subsection{Background}
Adversarial Attacks can be broken into two larger categories: black box attacks and white box attacks. With Black Box methods, the attacker does not have access to the model information such as the dataset or the architecture. With White Box attacks, it is assumed that an attacker will have access to the underlying model architecture and weights. Adversarial Attacks rely on a baseline knowledge of the underlying model including items like architecture, design, and dataset \cite{bhambri2019study}.  Black and White Box attacks are divided by the amount of incoming knowledge an attacker has about a system.

In the Cyber Security community, it is considered a huge advantage to know the underlying hardware, software, or network architecture when attacking a system \cite{uma2013survey}. Similarly, in Machine Learning, understanding the dataset, model architecture, or hardware it is running on can provide a large advantage to an attacker \cite{wang2019security}. This paper seeks to start exploring how can an adversarial agent can get access to this information without access to the model. This work covers a limited set of experiments across image and text classifiers to demonstrate the ability of this technique to find the underlying model design from simple probes like specific images in a dataset. 

\subsection{Challenges}
There are two types of Adversarial Attacks: universal and targeted. The attack is universal when they are able work across multiple architectures[citation needed] and the attack is targeted when it aims for a specific model design. Targeted attack's are more effective adversarial attacks \cite{chakraborty2018adversarial}. 

Our experiments are limited to target attacks on models or underlying weights learned from particular datasets. The focus is on detecting model information from downloadable models in popular libraries by probing text and image classifiers. If a machine learning team trains their model on a completely custom dataset, then this method would need to be extended to find in family or out of family examples. Finally, the key limitation is using publicly available datasets and architectures to build our training data for the classifier. For example, in the dogs dataset, a specific dog is chosen to produce a predictable output for the detector. The experiments assume datasets are not mixed or combined in any way.

\subsection{Contributions}
Adversarial Attack papers focus universal or targeted attacks to a particular model or architecture design. There is a prior step that is often overlooked in the adversarial attack process – discovering which attack is effective against a target. The contributions in this paper offer a process for discovering underlying model data for use in an adversarial attack system.




\section{Approach}
In this work, an approach is laid out to discover model attributes. Figure 1 shows the three steps involved: probe, collect, and detect. The following sections will cover the process design, experiment design, and attribution permutations.
\begin{figure}[t]
    \center
    \includegraphics[width=3.25in]{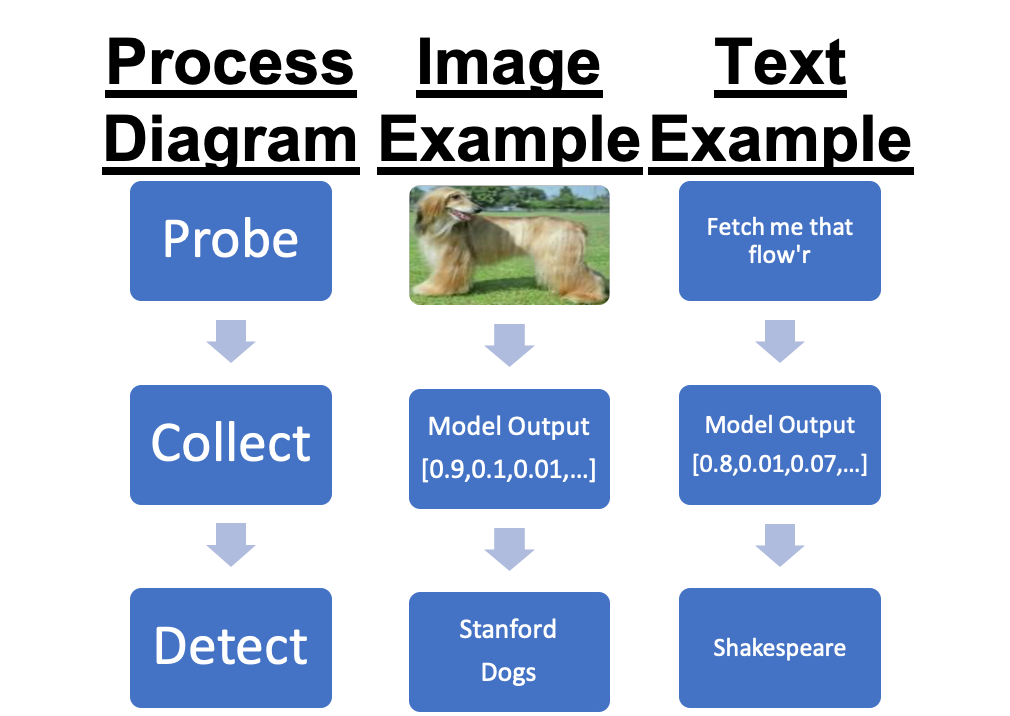}
    \caption{Approach to Discovering Model Attributes}
    \label{Figure 2}
\end{figure}
\subsection{Process Design}
This process is designed with repeatability in mind. Each probe selected is meant to represent a unique output for the underlying model and thus makes it easy for a classifier  to identify the attributes. The three stages of the process are probe, collect, and detect.
In the probe stage, the attacker selects an image or group of images that has unique features in comparison to other datasets. In Figure 1, an Afgan Hound Dog is chosen. This dog is not represented adequately in other datasets and represents a unique output for a classifier. The collect stage is where the attack system gathers the output of the targeted model and sends it to the detect stage. The initial experiments with this process use the full class/probability outputs. Last, in the detect stage, a classifier is trained to probe images and model outputs. With this classifier, it is possible to detect the model attributes with single image or text inputs. The results section will show initial results of this process.

\begin{figure*}[!t]
    \centering
    \includegraphics[width=7.25in]{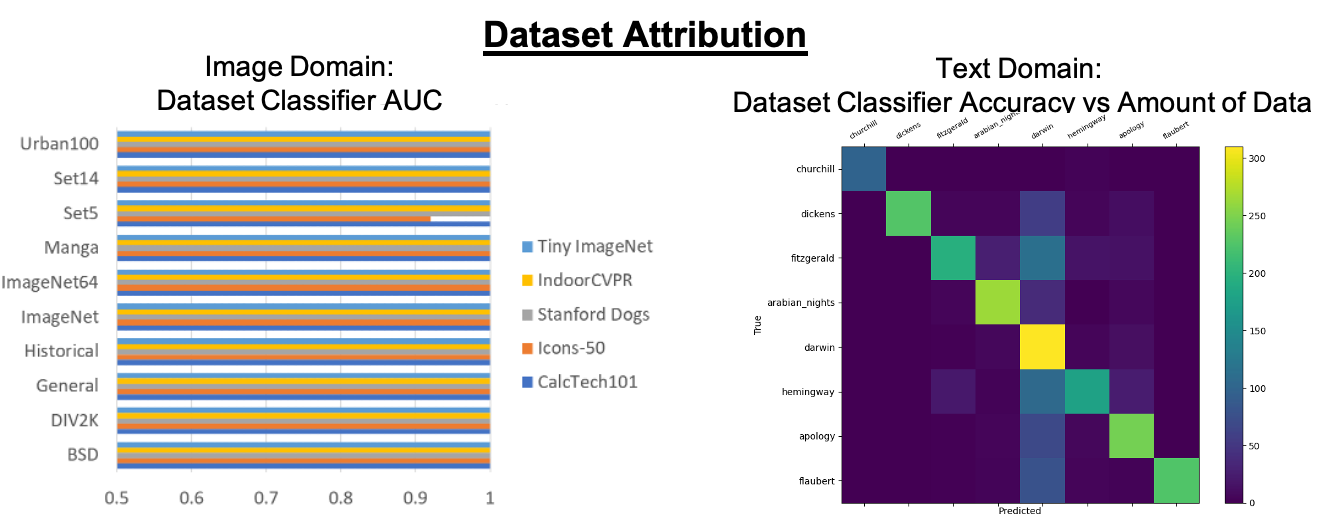}
    \caption{Results of Dataset Attribution Experiments}
    \label{Figure 1}
\end{figure*}

\subsection{Experiment Design}
The experiments will document image classification and natural language processing classification. Each domain in machine learning has common architectures and datasets that are used to create the base weights. The goal is to identify model attributes strictly from probing the final model. There are two types of model attributes that are covered here: architecture and dataset.

\subsubsection{Architecture Attribution}
There are hundreds of model architectures for each problem in a sub-domain. In the adversarial world, the potency of adversarial techniques relies on knowing the underlying architecture. With this approach of determining the architecture with only end user access to the model, the goal is to classify the classifier model design. 

In the image experiments, there are seven architectures explored. With Natural Language, GPT-2 \cite{radford2019language} is trained on downstream tasks using the two smaller models - 117M and 355M models. The classifier with text is attempting to distinguish these two text transformers from each other for multiple datasets.

\subsubsection{Dataset Attribution}
Each machine learning model uses training data to create base weights. Model fine tuning is defined as updating the model weights on a smaller, targeted dataset. This work shows the ability to detect the dataset that the base weights were derived from, even when a model has been fine tuned on another dataset.

In the image space, ten separate datasets are explored. The datasets are taken from super resolution datasets alongside a few popular image datasets. Similarly, on the text side,  recognizable datasets are taken from popular authors to show how each model learns and generates different underlying language models for a downstream task. The dataset attribution in Natural Language Processing (NLP) can be a more difficult task as permutations with language classification become intractable quickly (example: misspellings and colloquialisms). 

\section{Evaluation}
 Image and Natural Language Processing experiments are conducted with the intent of classifying the underlying model attributes. The image experiments explore pretrained image classifiers available in Keras while training on datasets publicly available. In Natural Language Processing, each experiment uses readily available datasets in the HuggingFace packages. Due to the computational complexity of training transformers, there were only two models explored of the popular GPT-2 model.

\subsection{Dataset Attribution Results}
Dataset attribution [Figure \ref{Figure 1}] relies on the ability of our classifier to access the resulting feature vector in the image space and the resulting text output of the transformer. All datasets are downloaded and deployed from public repositories.

\subsubsection{Determining the Dataset Type from a Single Image}
The following datasets were used with random 50 class subsets to fine tune MobileNetV2: [CalTech101, Icons-50, indoorCVPR, Stanford Dogs Dataset, tiny-imagenet]. After training, inference was applied to the datasets: [General, BSD100, DIV2K, Set5, Set14, Urban100, ImageNet64 (16k subset), ImageNet (6k subset), Manga, Historical] \cite{preet2020vision}. The inference vector output was captured alongside the fine tuned model which was used. A classifier is trained on each inference dataset to attempt to decipher and predict which fine tuned model the model output came from. The results are surprisingly good. For any of the given datasets used, it is possible to predict with an AP of greater than .99 for any of the fine tuned models with sklearn's standard Random Forest model.

\subsubsection{Determining the Dataset Type from a Single Text Input}
The following datasets are trained with a GPT-2 small model: Churchhill, Dickens, Fitzgerald, Arabian Nights, Darwin, Hemmingway, Apology, and Flaubert. The GPT-2 transformer models are trained using the HuggingFace repository for NLP research \cite{Wolf2019HuggingFacesTS}. With the dataset classifier using a Bert Large model, it is possible to achieve 81\% classification accuracy against each of these authors \cite{florio2020time}. In order to show the shortcomings of the current classifier, confusion matrix for the classifier appears in Figure 2. Overall, the model is able to clearly classify each of the authors. Notably in this matrix, there is some difficulty in identifying Apology and Darwin. This is likely due to how the transformer captures an author's style, relative to prose structure including whitespace and formatting.

\subsection{Architecture Attribution Results}
The next step is to classify the pretrained architectures [Figure \ref{Figure 2}] used in image classification and text generation. By applying different pretrained architectures onto different datasets, it is possible to learn to tell the different pretrained models apart. Once there is a classifier built from this dataset, then it can be used to probe the original model with selected image or text inputs. The same process is also used to attempt to classify the architecture of a model after fine tuning. The idea is to take a pretrained model and fine tune it on different datasets. We can then proceed to get outputs on a test set of images to probe and classify the underlying architecture.

\subsubsection{Determining the Architecture Type from a Single Image}
There are 10 datasets used in these experiments: [General, BSD100, DIV2K, Set5, Set14, Urban100, ImageNet64 (16k subset), ImageNet (6k subset), Manga, Historical] and the 7 following pretrained networks: [MobileNetV2, NASNetMobile, DenseNet121, ResNet50, DenseNet201, Xception, InceptionV3]. All pretrained models are from keras.applications which are pretrained on 1k classes of ImageNet \cite{chollet2015keras}. Inference is performed on each of the datasets for each model and the 1000-dimension output was captured. Then, for each dataset, the model's output was collected alongside the name of the model used. This captures the difference in each of the feature outputs and allows us to use a classifier that predicts from the inference results onto the model on which it came from.

For each dataset, sklearn's standard Random Forest model was able predict with an Average Precision (AP) of .84 for the worst-case scenario of five Set5 images. All other classifiers had an AP of .99. Therefore, given at least five images from a dataset and a sample image classification prediction, it is possible to classify a pretrained model.

\subsubsection{Determining the Architecture Type from a Text Sample}
Architecture Attribution with text transformers can be harder due to hardware and time limitations. For instance, recent research is focused on speeding up training of transformers by 10-40\% to make the retraining of these models more approachable \cite{notin2020sliceout}. Our experiments trained two separate models, GPT-2-small and DistilBert on the same dataset \cite{sanh2019distilbert}.  This experiment used the Wiki Text language modeling dataset  - a benchmark based on verified Wikipedia articles that provides fast and repeatable training results for many transformer architectures \cite{wikitext}. The experiments in [Figure \ref{Figure 2}] show that the accuracy of a 20,000 sample trained classified with our five probes will only show moderate predictability at 60\% accuracy for simple classifiers. The text probes were chosen as basic, approachable probes and not targeted to specific data in the models yet: 'Hello', '2+2', 'A', and 'Mario'. Given the Wikipedia input dataset, these probes should provide a diverse but distinguishable output. The results below demonstrate further honing of the target probes will lead to improved accuracy.

\subsection{Limitations}
Each of these experiments focused on demonstrating how distinguishable an model attributes are in the image and text domain. As architectures are modified, it may only be possible to find the most similar architecture to the ones in the training set. For instance, different model architectures may perform similarly when trained on the same datasets. The experiments outlined in this paper do not mix datasets or compare architecture similarities. These experiments utilized classifiers that are able to be directly downloaded from Keras or HuggingFace with no additional installations necessary. In the same way, each of the datasets are utilized with zero modification or augmentations. The effects of augmentations, architecture modification, and dataset mixing are left for a future work. 
\begin{figure*}[!t]
    \centering
    \includegraphics[width=7.25in]{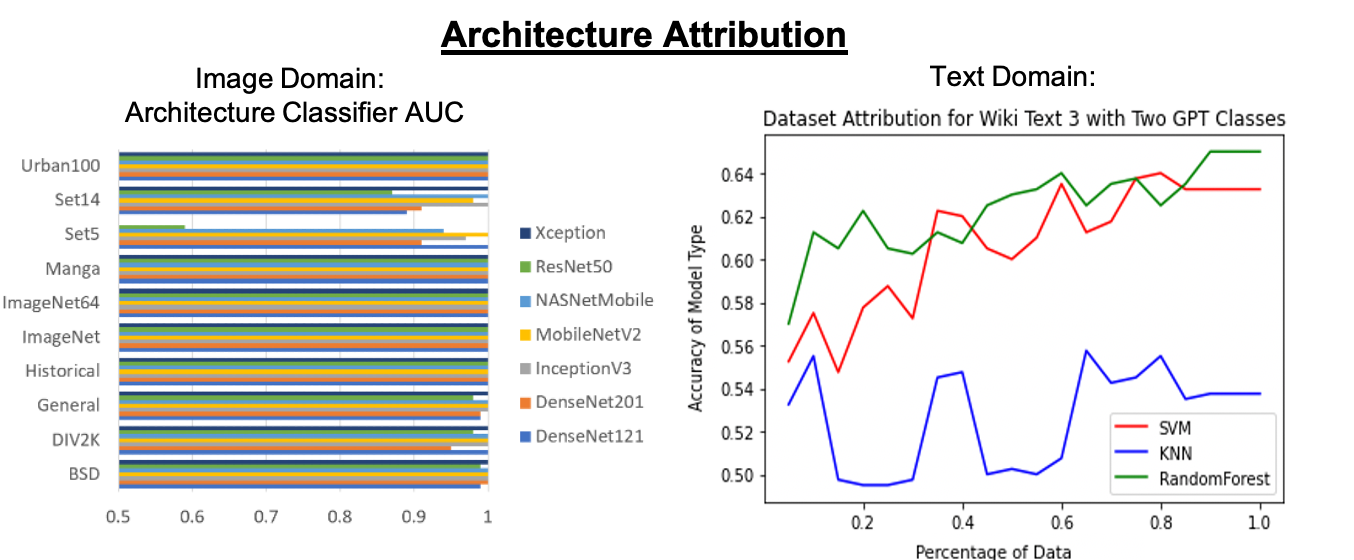}
    \caption{Results of Architecture Attribution Experiments}
    \label{Figure 2}
\end{figure*}
\section{Future Work}
In the text space, the technique needs to be refined down to only use the output from the model at an API or pipeline level. It is possible to also use targeted attacks known to work against particular models to determine the underlying architecture. If an adversarial attack is tailored to work against a particular model (and is not universal), then that technique would not be effective against tangential architecture types. Additionally in the text space, future work will focus on determining the number of samples needed to improve the efficacy of classifying the underlying model information. Larger datasets, like modern transformers that are trained on billions of documents, will require further investigation \cite{li2020train}. Future experiments would utilize larger versions of the GPT-2 model and even the recently released GPT-3 \cite{brown2020language}.

In the image space, further work with predicting families of classifier based on how similar their outputs are will allow us to group architectures for adversarial attacks. This will close the gap between a targeted attack on one model and a universal attack on all models. Specifically in the classifier space, the prevalence on transfer learning leaves the machine learning community at heavy risk of compromise.

\section{Conclusion}
The goal of this work is to demonstrate the ability to discover model attributes from simple input image or text probes. In the image space, learning the fingerprint of a model is achievable with modern classifiers. This process to architecture and dataset discovery  reaches high AP numbers with minimal training. In the text domain, classifying the underlying model architecture is harder with a single text sample.  For trained datasets, the results in the text domain showed that datasets with clear stylistic cues are distinguishable from each other. For the sample experiments, model attributes are discoverable with simple input probes.

\section*{Acknowledgment}
The authors would like to thank Auburn University and the PeopleTec Technical Fellows program for encouragement and project assistance.

\ifCLASSOPTIONcaptionsoff
  \newpage
\fi



%

\bibliographystyle{./bibtex/IEEEtran}
\bibliography{./IEEEexample}
\end{document}